\pdfoutput=1

\documentclass[11pt]{article}

\usepackage{acl}

\usepackage{times}
\usepackage{latexsym}

\usepackage[T1]{fontenc}

\usepackage[utf8]{inputenc}

\usepackage{microtype}

\usepackage{inconsolata}

\usepackage{graphicx}
\usepackage{amsmath} 
\usepackage{amsmath}
\usepackage{amssymb}
\usepackage{mathtools}
\usepackage{amsthm}
\usepackage[textsize=tiny]{todonotes}
\usepackage{enumitem}
\usepackage{booktabs}
\usepackage{pifont}
\usepackage{makecell}
\usepackage{tikz}
\usepackage{adjustbox}    
\usepackage{xcolor}
\usepackage{float}

\newif\ifshowblue
\showbluefalse   

\newcommand{\blueadd}[1]{%
  \ifshowblue
    \textcolor{blue}{#1}%
  \else
    #1%
  \fi
}

\newcommand*\circled[1]{\tikz[baseline=(char.base)]{
            \node[shape=circle,draw,inner sep=2pt] (char) {#1};}}

\newcommand{\cmark}{\ding{51}}  
\newcommand{\xmark}{\ding{55}}  

\title{ReviewEval: An Evaluation Framework for AI-Generated Reviews}

\author{
 \textbf{Madhav Krishan Garg\textsuperscript{1}},
 \textbf{Tejash Prasad\textsuperscript{1}},
 \textbf{Tanmay Singhal\textsuperscript{1}},
 \textbf{Chhavi Kirtani\textsuperscript{1}},
\\
 \textbf{Murari Mandal\textsuperscript{2}},
 \textbf{Dhruv Kumar\textsuperscript{1, 3}},
\\
 \textsuperscript{1}IIIT Delhi,
 \textsuperscript{2}KIIT Bhubaneswar,
 \textsuperscript{3}BITS Pilani,
\\
 \small{
   \textbf{Correspondence:} \href{mailto:dhruv.kumar@pilani.bits-pilani.ac.in}{dhruv.kumar@pilani.bits-pilani.ac.in}
 }
}

\begin{document}
\maketitle
\begin{abstract}

The escalating volume of academic research, coupled with a shortage of qualified reviewers, necessitates innovative approaches to peer review. In this work, 
we propose: 
\ding{182} ReviewEval, a comprehensive evaluation framework for AI‐generated reviews that measures alignment with human assessments, verifies factual accuracy, assesses analytical depth, identifies degree of constructiveness and adherence to reviewer guidelines; and \ding{183} 
ReviewAgent, an LLM‐based review generation agent featuring a novel alignment mechanism to tailor feedback to target conferences and journals, along with a self‐refinement loop that iteratively optimizes its intermediate outputs and an external improvement loop using ReviewEval to improve upon the final reviews. 
ReviewAgent improves actionable insights by 6.78\% and 47.62\% over existing AI baselines and expert reviews respectively. Further, it boosts analytical depth by 3.97\% and 12.73\%, enhances adherence to guidelines by 10.11\% and 47.26\% respectively. 
This paper establishes essential metrics for AI‐based peer review and substantially enhances the reliability and impact of AI‐generated reviews in academic research.

\end{abstract}

\section{Introduction}
The rapid growth of academic research, coupled with a shortage of qualified reviewers, has created an urgent need for scalable and high-quality peer review processes \cite{petrescu2022evolving, schulz2022future, checco2021ai}. This need has led to a growing interest in leveraging large language models (LLMs) to automate and enhance various aspects of the peer review process~\cite{robertson2023gpt4, liu2023reviewergpt}.\par

Although LLMs have shown remarkable potential in automating various natural language processing tasks, their effectiveness in serving as reliable and consistent paper reviewers remains a significant challenge.
The academic community is already experimenting with AI-assisted reviews; for instance, 15.8\% of ICLR 2024 reviews involved AI assistance~\cite{latona2024ai}. Despite increasing LLM adoption, concerns about reliability and fairness remain.

\begin{figure}[t]
    \centering
    \tiny
    \includegraphics[width=0.45\textwidth]{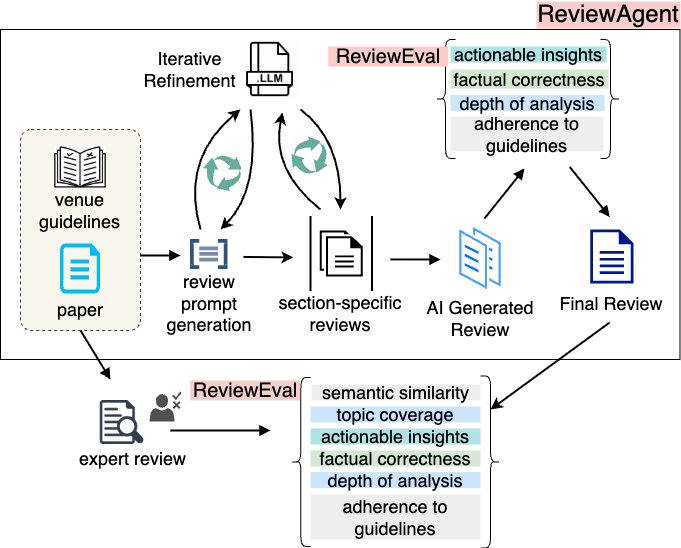}  
    \caption{ReviewEval and ReviewAgent: Given a paper and the associated conference or journal guidelines, ReviewAgent generates AI-based reviews and evaluates them along multiple dimensions using ReviewEval.}
    \label{fig:overview}
    \vspace{-2em}
\end{figure}

Specifically, papers reviewed by AI have been perceived to gain an unfair advantage, leading to questions about the integrity of such evaluations. Consequently, research into robust automated review generation systems is crucial, necessitating rigorous evaluation of AI generated reviews to address key challenges.
\cite{zhou2024llm} analyze commercial models like GPT-3.5 and GPT-4~\cite{achiam2023gpt} as paper reviewers, identifying limitations such as hallucinations, incomplete understanding, and insufficient critical feedback compared to humans.

Existing research on evaluation metrics for AI-generated research paper reviews remains limited. 
For instance,~\cite{d2024marg} proposed a GPT-4-based automated metric for evaluating approximate matches between AI-generated and human-written reviews. However, their method's iterative reliance on GPT-4 creates a black-box evaluation, limiting transparency and raising reliability concerns.
Similarly, \cite{zhou2024llm} investigated the aspect coverage and similarity between AI and human reviews through a blend of automatic metrics and manual analysis. 
But their approach overlooks other critical dimensions where AI reviews may underperform, as highlighted in Figure~\ref{fig:challenges}.\par




Addressing the above mentioned limitations and gaps, 
we propose ReviewEval, a comprehensive evaluation framework for assessing the quality of AI-generated research paper reviews. ReviewEval targets five key dimensions (see Figure~\ref{fig:challenges}): \ding{182} \textit{Comparison with Human Reviews}: Evaluates topic coverage and semantic similarity to measure the alignment between AI-generated and human-written feedback. \ding{183}
\textit{Factual Accuracy}: Detects factual errors, including misinterpretations, incorrect claims, and hallucinated information. \ding{184} \textit{Analytical Depth}: Assesses whether the AI’s critique transcends generic commentary to offer in-depth, meaningful feedback. \ding{185} \textit{Actionable Insights}: Measures the ability of the AI to provide constructive suggestions for improving the paper. \ding{186} \textit{Adherence to Reviewer guidelines}: Quantifies the degree to which a review conforms to the evaluation criteria outlined by the target conference.\par

Recent studies \cite{bauchner2024use, biswas2024ai} underscore the growing importance of aligning reviews with conference-specific evaluation criteria, especially as many venues now require adherence to detailed reporting guidelines. To address this, 
we introduce ReviewAgent, an AI reviewer with three key features: \ding{182} Conference-specific review alignment, dynamically adapting reviews to each venue's unique evaluation criteria; \ding{183} Self-refinement loop, iteratively optimizing prompts and intermediate outputs for deeper analytical feedback; \blueadd{\ding{184} External improvement loop, systematically enhancing review quality using evaluation metrics provided by ReviewEval.}

\blueadd{Using ReviewEval, we extensively evaluate ReviewAgent against two baseline models, AI-Scientist ~\cite{lu2024ai} and MARG ~\cite{d2024marg}, benchmarking all systems against expert-written reviews treated as the gold standard. The experiments are performed on a balanced and topically diverse dataset of 120 papers sampled from NeurIPS, ICLR, and UAI, including domains such as computer science, biology, social science, finance, physics, and engineering.}

\blueadd{Our results show that ReviewAgent improves actionable insights by 6.78\% and 47.62\% over existing AI baselines and expert reviews respectively. Further, it boosts analytical depth by 3.97\% and 12.73\%, enhances adherence to guidelines by 10.11\% and 47.26\% respectively.}



In summary, the paper makes the following research contributions: 
\textbf{(1) ReviewEval}: a multi-dimensional framework for evaluating research paper reviews (\S \ref{sec:review_eval}). \textbf{(2) ReviewAgent}: an adaptive AI reviewer which utilizes ReviewEval along with novel conference alignment and iterative self-improvement (\S \ref{sec:review_agent}). \textbf{(3) Comprehensive evaluation} of ReviewAgent against state-of-the-art baselines using ReviewEval (\S \ref{results-section}).

\section{Related Work}
\noindent\textbf{AI-based scientific discovery.} 
Early attempts to automate research include 
AutoML that optimize hyperparameters and architectures~\cite{hutter2019automated, he2021automl} and AI-driven discovery in materials science and synthetic biology~\cite{merchant2023ai, hayes2024simulating}. However, these methods remain largely dependent on human-defined search spaces and predefined evaluation metrics, limiting their potential for open-ended discovery. Recent works~\cite{lu2024ai} aim to automate the entire research cycle, encompassing ideation, experimentation, manuscript generation, and peer review, thus pushing the boundaries of AI-driven scientific discovery.\par

\begin{figure}[t]
    \centering
    \includegraphics[width=0.5\textwidth, trim=0cm 0cm 0cm 0cm, clip]{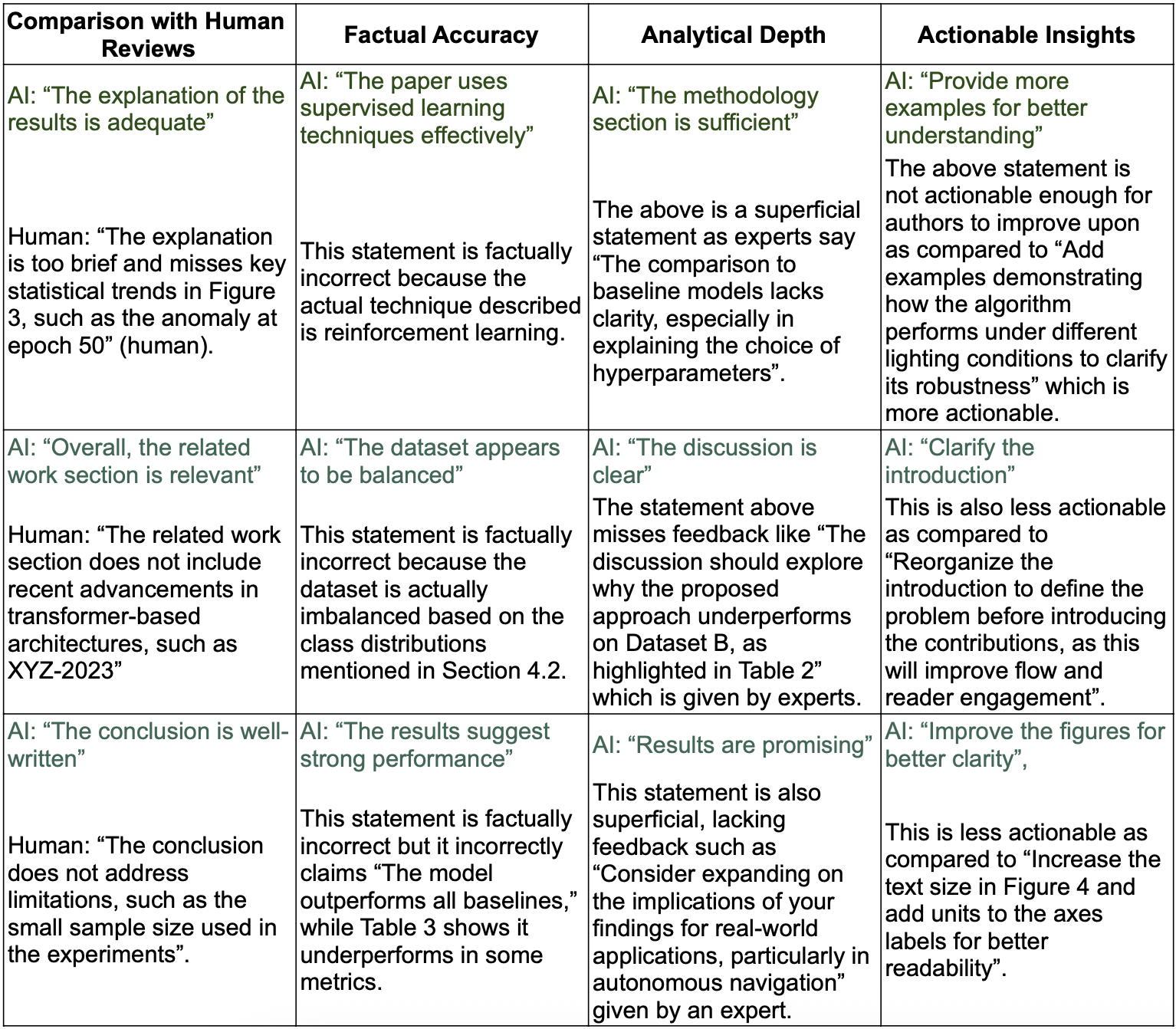} 
    \caption{\small Illustrative examples highlighting key challenges in AI-generated research paper reviews that motivate our proposed evaluation metrics: Column 1 shows semantic and topical divergences from human reviews, supporting the Human Comparison metric; Column 2 presents factual inaccuracies or hallucinations, motivating the Factual Accuracy metric; Column 3 illustrates limited analytical reasoning, justifying the Analytical Depth metric; Column 4 reveals a lack of specific, actionable suggestions, underscoring the need for the Actionable Insights metric.}
    \vspace{-2em}
    \label{fig:challenges}
\end{figure}

\noindent\textbf{AI-based peer-review.}
Existing work has looked at scoring and improving research papers in a variety of ways such as statistical reporting inconsistencies ~\cite{nuijten_statcheck_2020}, recommending citations ~\cite{ALI2020113790} and predicting review scores ~\cite{basuki2022review, bharti_peerrec_2024}. More recently, LLM-based approaches have been used to generate peer reviews ~\cite{robertson2023gpt4, liu2023reviewergpt, d2024marg, lu2024ai, liang_can_2023}.

~\cite{lu2024ai} employ LLMs to autonomously conduct the research pipeline, including peer review. It follows a structured three-stage review process: paper understanding, criterion-based evaluation (aligned with NeurIPS and ICLR guidelines), and final synthesis: assigning scores to key aspects like novelty, clarity, and significance. 
MARG~\cite{d2024marg} introduces a multi-agent framework where worker agents review sections, expert agents assess specific aspects, and a leader agent synthesizes feedback. Using BERTScore~\cite{bert-score} and GPT-4-based evaluation, MARG-S improves feedback quality, reducing generic comments and increasing helpful feedback per paper. 
These studies highlight the AI's potential to enhance peer review through structured automation and multi-agent collaboration.\par

\noindent\textbf{Evaluation framework for AI-based peer-review.} There has been limited research on developing evaluation frameworks for evaluating the quality of LLM generated paper reviews.~\cite{zhou2024llm} evaluated GPT models for research paper reviewing across 3 tasks: aspect score prediction, review generation, and review-revision MCQ answering. Their evaluation framework comprised aspect coverage (originality, soundness, substance, replicability, etc.), ROUGE (lexical overlap), BERTScore (semantic similarity), and BLANC (informativeness), alongside manual analysis. Results showed LLMs overemphasized positive feedback, lacked critical depth, and neglected substance and clarity, despite high lexical similarity to human reviews. \cite{d2024marg} introduced an automated evaluation framework for AI-generated reviews, quantifying similarity to human reviews via recall, precision, and Jaccard index. 
Recall measures the fraction of real-reviewer comments with at least one AI match, precision quantifies AI comments aligned with human reviews, and Jaccard index evaluates the intersection-over-union of aligned comments.\par

Existing evaluation metrics predominantly emphasize the similarity between AI-generated and human reviews, overlooking other crucial parameters. Moreover, their heavy reliance on LLMs for end-to-end evaluation results in a black-box system with limited transparency. In contrast, our framework introduces more interpretable evaluation metrics for AI-generated reviews (see Figure~\ref{fig:challenges} and Table~\ref{tab:evaluation_comparison}), effectively addressing these shortcomings.

\noindent\textbf{Iterative Refinement of Large Language Models.}
\blueadd{Recent studies highlight the benefits of iterative feedback-driven refinement in improving LLM outputs: Self-Refine \cite{madaan2023selfrefine} uses self-generated feedback without additional training, LLMRefine \cite{xu2024llmrefine} employs a learned feedback model with simulated annealing, and ProMiSe \cite{hu2024teaching} leverages external proxy metrics, collectively enhancing factuality, coherence, and performance in tasks like document-grounded QA and dialog generation.}

\begin{table}[t]
    \centering
    \tiny
    \resizebox{\columnwidth}{!}{%
    \begin{tabular}{c c c c c c}
        \toprule
        Method & depth & factual & topic & actionable\\
        & analysis & correctness & coverage & insight \\  \midrule
        \cite{d2024marg} & \xmark & \xmark & \xmark & \cmark\\
        \midrule
        \cite{zhou2024llm} & \xmark & \xmark & \xmark & \xmark\\
        \midrule
        ReviewEval (ours) & \cmark & \cmark & \cmark & \cmark\\
        \bottomrule
    \end{tabular}
    }
    \caption{Existing metrics focus on AI-human review similarity but overlook other key aspects; our framework fills these gaps with more interpretable metrics.}
    \vspace{-20pt}
    \label{tab:evaluation_comparison}
\end{table}

\section{ReviewEval}\label{sec:review_eval}
We introduce ReviewEval, an evaluation framework grounded in the LLM-as-a-Judge paradigm \cite{gu2025surveyllmasajudge}. 
In ReviewEval, each review is evaluated on several key parameters to assess the overall quality of the generated feedback. To ensure consistency and reliability, all evaluations for a given metric were performed by LLMs of the same specification and version, \blueadd{minimizing variability from model differences and enabling robust, fair, and unbiased comparisons.} The overview of the proposed framework (ReviewEval + ReviewAgent) is presented in Figure~\ref{fig:overview}.\par

\subsection{Comparison with Expert Reviews}
We compare the reviews generated by the LLM based reviewer with expert human reviews. Our primary goal is to gauge how well the AI system replicate expert-level critique. The evaluation is conducted along the following dimensions:

\textbf{Semantic similarity.} 
To assess the alignment between AI-generated and expert reviews, we embed each review \( R \) into a vector representations using a text embedding model \cite{mikolov2013efficientestimationwordrepresentations}. The semantic similarity between an AI-generated review \( R_{\text{AI}} \) and an expert review \( R_{\text{Expt}} \) is measured using cosine similarity:
\vspace{-1em}
\begin{equation}
S_{\text{sem}}(R_{\text{AI}}, R_{\text{Expt}}) = \frac{e(R_{\text{AI}}) \cdot e(R_{\text{Expt}})}{\|e(R_{\text{AI}})\| \, \|e(R_{\text{Expt}})\|}
\end{equation}
\vspace{-0.1em}
where \( e(R) \) denotes the embedding of review \( R \). A higher cosine similarity indicates a stronger alignment between the AI-generated and expert reviews.\par


\textbf{Topic Coverage.} We evaluate topic coverage to determine how comprehensively AI-generated reviews address the breadth of topics present in expert reviews. Our approach comprises three steps: \ding{182} \textit{Topic extraction:} Each review \( R \) (either AI-generated or expert) is decomposed into a set of topics by an LLM: $T_R = \{ t_1, t_2, \dots, t_n \},$ where each topic \( t_i \) is represented by a sentence that captures its core content and context. \ding{183} \textit{Topic similarity:} Let $T_{\text{AI}} = \{ t_1, t_2, \dots, t_m \}$ and  $T_{\text{Expt}} = \{ t'_1, t'_2, \dots, t'_n \}$
denote the topics extracted from the AI and expert reviews, respectively. We define a topic similarity function \( \text{TS}(t_i, t'_j) \) that an LLM assigns on a discrete scale:\\
\(\text{TS}(t_i, t'_j) = 3 \cdot \mathbb{I}\{ t_i \sim_{\text{strong}} t'_j \} + 2 \cdot \mathbb{I}\{ t_i \sim_{\text{moderate}} t'_j \} + 1 \cdot \mathbb{I}\{ t_i \sim_{\text{weak}} t'_j \}\), where $\mathbb{I}$ is the indicator function, $t_i \sim_{\text{strong}} t'_j$, $t_i \sim_{\text{moderate}} t'_j$ , $t_i \sim_{\text{weak}} t'_j$ denote substantial, moderate, and minimal overlap in concepts, respectively. All the conditions are mutually exclusive. \blueadd{We set a similarity threshold \( \tau = 2 \) so that a topic in AI-generated review having atleast moderate similarity to a topic in expert review is considered aligned.} \ding{184} \textit{Coverage ratio:} For each AI-generated review, we construct a topic similarity matrix \( S \) where each element $S[i, j] = \text{TS}(t_i, t'_j)$ represents the similarity between topic \( t_i \) from \( T_{\text{AI}} \) and topic \( t'_j \) from \( T_{\text{Expt}} \). The topic coverage ratio is defined as:
\vspace{-0.8em}
\begin{equation}
S_{\text{coverage}} = \frac{1}{n} \sum_{j=1}^{n} \mathbb{I}\left(\max_{i=1,\dots,m} S[i, j] \geq \tau\right),
\end{equation}
where \( \mathbb{I}(\cdot) \) is the indicator function, and \( n = |T_{\text{Expt}}| \) is the total number of topics extracted from the expert review.

\subsection{Factual Correctness}
To address hallucinations and factual inaccuracies in LLM-generated reviews, we introduce an automated pipeline that simulates the conference rebuttal process, allowing evidence-based validation of reviewer claims. By automating both the question-generation and rebuttal phases, our system produces a robust factual correctness evaluation. The pipeline consists of the following steps:\par

\textit{\textbf{Step 1:} Question generation.}  
Each review \(R\) is converted by an LLM into a structured verification question \(Q\) that captures its central critique (Example in Appendix \ref{factual_correctness_examples}). 

\textit{\textbf{Step 2:} Query decomposition.}  
A dedicated LLM-based decomposition component splits \(Q\) into a set of sub-questions \(\{q_1, \ldots, q_n\}\), isolating distinct aspects of the claim for fine-grained analysis. (Implementation and Example in Appendix \ref{factual_correctness_examples}.)

\textit{\textbf{Step 3:} Retrieval-augmented synthesis.}  
For each sub-question \(q_i\), we:
\circled{a} \blueadd{retrieve relevant text segments (\(\approx 1000\) tokens) via semantic similarity search over contextual embeddings;}
\circled{b} \blueadd{extract the corresponding parent sections (\(\approx 4000\) tokens) to provide broader context. Both parent and child document splitters use a 10\% overlap to enhance context retention across chunk boundaries;}
\circled{c} generate a focused answer \(A_i\) to \(q_i\) using the retrieved context.
The individual answers are then aggregated into a unified response \(A_Q\) using an LLM addressing the original question \(Q\).\par

\textit{\textbf{Step 4:} Automated rebuttal generation.}  
The aggregated evidence \(A_Q\) serves as the basis for an evidence-based rebuttal \(R_b\), which systematically supports or counters each claim in \(R\) by citing specific sections of the paper.\par

\textit{\textbf{Step 5:} Factual correctness evaluation.}  
We compare \(R\) against \(R_b\) to determine whether each claim is substantiated:
\vspace{-0.5em}
\[
\mathcal{V} = \left\{
\begin{array}{ll}
1\!: & \text{all claims in } R \text{ are supported by } R_b; \\
0\!: & \text{otherwise}
\end{array}
\right.
\]
We then compute an overall factual correctness score:
\vspace{-1em}
\begin{equation}
S_{\text{factual}} = \frac{1}{|Q|} \sum_{i=1}^{|Q|} \mathbb{I}\bigl(\mathcal{V}_i = \text{1} \bigr),
\end{equation}
\vspace{-2em}


\subsection{Constructiveness}
We assess review constructiveness by quantifying the presence and quality of actionable insights in AI-generated reviews relative to expert feedback. Our framework begins by extracting key actionable components from each review using an LLM with few-shot examples. Specifically, we identify the following  insights: \blueadd{(i) \emph{criticism points} (\(CP\)), which capture highlighted flaws or shortcomings in the paper’s content, clarity, novelty, and execution; (ii) \emph{methodological feedback} (\(MF\)), which encompasses detailed analysis of experimental design, techniques, and suggestions for methodological improvements; and (iii) \emph{suggestions for improvement} (\(SI\)), which consist of broader recommendations for enhancement such as additional experiments, alternative methodologies, or improved clarity.}\par

\blueadd{Once these components are extracted, each insight is evaluated along three dimensions: \textbf{specificity}, \textbf{feasibility}, and \textbf{implementation details}.}
\vspace{-0.5em}
\setlength{\itemsep}{0pt}\begin{itemize}[leftmargin=*]
    \item \blueadd{The \textbf{specificity score} $\sigma$ is defined as 1 if the insight is clear and unambiguous, referring to a particular aspect of the paper and including explicit examples.}
    \vspace{-0.8em}
    \item The \textbf{feasibility score} \blueadd{$\phi$ = 1 when the recommendation can realistically be implemented within the research context, such as available data, reasonable technical effort, or domain constraints.}
    \vspace{-2em}
    \item The \textbf{implementation detail score} \blueadd{$\delta$ = 1 if the feedback provides actionable steps, specific techniques, or references for improvement.}
\end{itemize}
\vspace{-0.5em}
Examples explaining the three dimensions in Appendix  \ref{constructiveness_example}.

 The overall actionability score for an individual insight is then computed as $S_{\text{act},i} = \sigma_i + \phi_i + \zeta_i,$ with an insight considered actionable if \(S_{\text{act},i} > 1\) (having atleast 2 of the three qualities mentioned above). Finally, we quantify the overall constructiveness of a review by calculating the percentage of actionable insights:
 \vspace{-1em}
\begin{equation}
 S_{\text{act}} = \frac{1}{N} \sum_{i=1}^{N} \mathbb{I}\big(S_{\text{act},i} > 1\big) \times 100,   
\end{equation}
\vspace{-0.1em}
where \(N\) is the total number of extracted insights and \(\mathbb{I}(\cdot)\) denotes the indicator function. This metric provides a quantitative measure of how  a review offers guidance for improving the work.

\subsection{Depth of Analysis}
To assess whether a review provides a comprehensive, critical evaluation rather than a superficial commentary, we measure the depth of analysis in AI-generated reviews. 
Each review is evaluated by multiple LLMs, which assign scores for each dimension, \(m_i\) (\(i \in \{1,2,3,4,5\}\)), with scores \(S_i \in \{0,1,2,3\}\). We define these as follows:\par

\blueadd{\textbf{Comparison with existing literature (\(m_1\)):} Assesses whether the review critically examines the paper's alignment with prior work, acknowledging relevant studies and identifying omissions. The scoring rubric is:\\$S_1 = \{3\!:\text{ thorough and critical comparison};\;2\!:\text{ meaningful but shallow};\;1\!:\text{ vague};\;0\!:\text{ not present}\}$ }


\blueadd{\textbf{Logical gaps identified (\(m_2\)):} Evaluates the review's ability to detect unsupported claims, reasoning flaws, and to offer constructive suggestions:\\ $S_2 = \{3\!:\text{ clear gaps with suggestions};\;2\!:\text{ some gaps, unclear recommendations};\;1\!:\text{ vague gaps, no solutions};\;0\!:\text{ no gaps identified}\}$}

\blueadd{\textbf{Methodological scrutiny (\(m_3\)):} Measures the depth of critique regarding the paper’s methods, including evaluation of strengths, limitations, and improvement suggestions:\\$S_3 = \{3\!:\text{ thorough and actionable};\;2\!:\text{ meaningful but limited};\;1\!:\text{ vague};\;0\!:\text{ none}\}$}



\blueadd{\textbf{Results interpretation} $(m_4)$: Assesses depth of result discussion (biases, alternative explanations, implications).\\
$S_4=\{3\!:\text{ insightful};\;2\!:\text{ shallow};\;1\!:\text{ vague};\;0\!:\text{ none}\}$}

\blueadd{\textbf{Theoretical contribution (\(m_5\)):} Evaluates the assessment of the paper’s theoretical contributions, including its novelty and connections to broader frameworks:\\$
S_5 = \{3\!:\text{ thorough and insightful critique};\;2\!:\text{ reasonable but lacks depth};\;1\!:\text{ vague critique};\;0\!:\text{ no assessment}\}
$}

The overall depth of analysis score for a review is calculated as the average normalized score across all dimensions:
\( S_{\text{depth}} = \displaystyle\frac{\sum_{i=1}^{5} S_i}{15} \)

A higher \(S_{\text{depth}}\) indicates a more comprehensive and critical engagement with the manuscript.

\subsection{Adherence to Reviewer Guidelines}
To assess whether a review complies with established criteria, we evaluate its adherence to guidelines set by the venue. 

Our approach begins by extracting the criteria \(C\) from the guidelines \(G\). These criteria fall into two broad categories: \ding{182} \emph{subjective} criteria, which involve qualitative judgments (e.g., clarity, constructive feedback), and \ding{183} \emph{objective} criteria, which are quantifiable (e.g., following a prescribed rating scale). For each review \(R\), every extracted criterion \(C_i\) is scored on a 0-3 scale using a dedicated LLM with dynamically generated prompts that include few-shot examples for contextual calibration. For subjective criteria, the score is defined as:\\ \blueadd{$
S_i = \{3\!:\text{ strong, detailed alignment};\;2\!:\text{ mostly aligned, minor issues};\;1\!:\text{ incomplete or inaccurate};\;0\!:\text{ no alignment}\}
$}

For objective criteria, the scoring is binary:\\ \blueadd{$
S_i = \{3\!:\text{ adheres to scale and structure};\;0\!:\text{ otherwise}\}
$}

The overall adherence score is then computed as \( S_{\text{adherence}} = \frac{\sum_{i=1}^{2} S_i}{6} \).

This score quantifies how well the review adheres to the prescribed guidelines.

\section{ReviewAgent}\label{sec:review_agent}

\blueadd{To ensure AI-based peer reviews are thorough across the multi-dimensional metrics defined in ReviewEval, we introduce ReviewAgent, a framework that aligns evaluations with conference-specific reviewing guidelines while operating effectively across those dimensions.}

\subsection{Conference-Specific Review Alignment}

To tailor the review process to conference-specific guidelines, we first retrieve them from the target conference's official reviewing website and pre-process them using LLM.
Each review guideline \(g_i\) is then converted into a step-by-step instructional prompt via LLM: \(P_i = \text{GeneratePrompt}(g_i)\),
where \(P_i\) denotes the prompt corresponding to guideline \(g_i\). Since some criteria apply to multiple sections of a research paper, the prompts are dynamically mapped to the relevant sections. This is achieved using a mapping function:
\(S_j = \mathcal{M}(P_i)\),
where \(S_j\) is the set of paper sections associated with prompt \(P_i\). Notably, \(\mathcal{M}\) is a one-to-many mapping, i.e., 
\(\mathcal{M}: P \to \mathcal{P}(S)\),
with \(\mathcal{P}(S)\) denoting the power set of all sections. By conducting reviews on a section-wise basis, our framework enhances processing efficiency and allows for independent evaluation of each section prior to aggregating the final review. 
(Prompts provided in Appendix \ref{prompts-used})

\subsection{Iterative Refinement Loop}
\label{relfection_loop}
\blueadd{To enhance the quality and completeness of our review prompts and the section-wise reviews generated, we employ an iterative refinement process using a Supervisor LLM. Starting with an initial set of prompts generated from the reviewing guidelines, the Supervisor LLM evaluates each prompt in conjunction with its corresponding guideline, serving as the problem statement, and provides targeted feedback. This feedback addresses key aspects such as clarity 
logical consistency 
alignment with the guideline, and comprehensiveness. 
The feedback is then used to revise the prompt, and this iterative loop is repeated for a fixed number of iterations (one in our current implementation). The result is a final set of structured, high-quality review prompts that are well-aligned with the reviewing guidelines and optimized for the evaluation process. We employ the same mechanism to iteratively refine the section-wise reviews generated, ensuring each section benefits from targeted feedback and improvement.}

\subsection{Improvement using ReviewEval}
\label{iterative_loop}
\blueadd{To further improve the quality and consistency of our reviews, we use an automated improvement loop based on an ReviewEval. After a draft for review is generated for a paper, ReviewEval evaluates it across four important dimensions: (1) constructiveness,(2) depth of analysis, (3) factual accuracy, and (4) adherence to reviewer guidelines a score along with scores of each criteria in these metrics. These are then passed on to the agent, which consumes both the raw review data and the evaluation feedback to generate an enhanced, final review. In our current implementation this loop is run once.\\Notably, throughout system development and benchmarking we do conduct an AI vs Human analysis in ReviewEval to judge AI generated reviews against expert reviews taking them as gold standard. But for real world submissions where we do not have any existing reviewer comments, we solely depend on these four objective criteria.}

\begin{table*}[t]
\centering
\resizebox{\hsize}{!}{
\begin{tabular}{l c c c c c c}
\toprule
\textbf{Framework} & Actionable & Adherence to & Coverage & Semantic & Depth of & Factual \\
& insights & review guidelines & of topics & similarity & analysis & correctness \\
\midrule
Expert & $0.4457 \pm 0.1322$ & $0.5967 \pm 0.1369$ & N/A $\pm$ N/A & N/A $\pm$ N/A & $0.8772 \pm 0.1127$ & $0.9797 \pm 0.0400$ \\
\midrule
MARG-V1-Deepseek & $0.4915 \pm 0.1193$ & $0.4367 \pm 0.1441$ & $0.5315 \pm 0.2139$ & $0.7957 \pm 0.0710$ & $0.8022 \pm 0.1231$ & $0.9873 \pm 0.0283$ \\
MARG-V1-Qwen     & $\mathbf{0.6161 \pm 0.1098}$ & $0.5160 \pm 0.1411$ & $0.5123 \pm 0.2057$ & $0.7809 \pm 0.0818$ & $0.9098 \pm 0.0929$ & $0.9777 \pm 0.0388$ \\
\midrule
MARG-V2-Deepseek & $0.3702 \pm 0.1603$ & $0.5898 \pm 0.1376$ & $0.8694 \pm 0.1625$ & $0.7638 \pm 0.0918$ & $0.9122 \pm 0.0629$ & $\mathbf{0.9911 \pm 0.0305}$ \\
MARG-V2-Qwen     & $0.4874 \pm 0.1441$ & $0.6337 \pm 0.1418$ & $0.8341 \pm 0.1856$ & $0.7814 \pm 0.0809$ & $0.9429 \pm 0.0544$ & $0.9762 \pm 0.0565$ \\
\midrule
ReviewAgent-Deepseek & $0.3096 \pm 0.1438$ & $0.7584 \pm 0.1266$ & $0.7330 \pm 0.1709$ & $0.8016 \pm 0.0990$ & $0.8694 \pm 0.1239$ & $0.9870 \pm 0.0556$ \\
ReviewAgent-Qwen     & $0.5899 \pm 0.1278$ & $\mathbf{0.8545 \pm 0.1039}$ & $0.7351 \pm 0.1803$ & $0.7887 \pm 0.1089$ & $\mathbf{0.9733 \pm 0.0514}$ & $0.9793 \pm 0.0436$ \\
\midrule
AI-Scientist-Deepseek & $0.3164 \pm 0.1611$ & $0.5997 \pm 0.1077$ & $\mathbf{0.8913 \pm 0.1450}$ & $0.7876 \pm 0.1012$ & $0.7045 \pm 0.1320$ & $0.9743 \pm 0.0966$ \\
AI-Scientist-Qwen     & $0.4684 \pm 0.1641$ & $0.7105 \pm 0.1508$ & $0.7933 \pm 0.1696$ & $\mathbf{0.8088 \pm 0.0988}$ & $0.9010 \pm 0.0965$ & $0.9901 \pm 0.0330$ \\
\bottomrule
\end{tabular}
}
\vspace{-0.5em}
\caption{Evaluation of AI-generated reviews across six different metrics on FullCorpus-120 dataset. ReviewAgent is compared against MARG~\cite{d2024marg} and Sakana AI Scientist~\cite{lu2024ai}. The 'NA' for Expert indicates that the metric is calculated relative to the Expert rating, and therefore does not require a comparison}
\vspace{-0.5em}
\label{tab:all_results}
\end{table*}

\begin{table*}[t]
\centering
\resizebox{\hsize}{!}{
\begin{tabular}{l c c c c c c}
\toprule
\textbf{Framework} & Actionable & Adherence to & Coverage & Semantic & Depth of & Factual \\
& insights & review guidelines & of topics & similarity & analysis & correctness \\
\midrule
MARG-V1-3.5-Haiku      & $0.4262 \pm 0.1287$ & $0.4587 \pm 0.1004$ & $0.5096 \pm 0.1985$ & $0.4330 \pm 0.0684$ & $0.4689 \pm 0.1551$ & $0.9650 \pm 0.0711$ \\
MARG-V1-3.7-Sonnet     & $0.4725 \pm 0.1017$ & $0.5820 \pm 0.0854$ & $0.4355 \pm 0.2203$ & $0.5057 \pm 0.0782$ & $0.8244 \pm 0.0884$ & $0.9593 \pm 0.0673$ \\
MARG-V1-GPT4o         & $0.3762 \pm 0.1116$ & $0.5567 \pm 0.1159$ & $0.4870 \pm 0.1906$ & $0.5019 \pm 0.0705$ & $0.7178 \pm 0.1283$ & $0.9587 \pm 0.1089$ \\
MARG-V1-GPT4o-mini    & $0.4338 \pm 0.1139$ & $0.5433 \pm 0.0807$ & $0.5034 \pm 0.1684$ & $0.4974 \pm 0.0700$ & $0.7867 \pm 0.0882$ & $0.9877 \pm 0.0344$ \\
\midrule
MARG-V2-3.5-Haiku      & $0.2746 \pm 0.1394$ & $0.4400 \pm 0.1400$ & $\bf{0.9467 \pm 0.1383}$ & $0.7025 \pm 0.1100$ & $0.7311 \pm 0.0934$ & $0.9713 \pm 0.0801$ \\
MARG-V2-3.7-Sonnet     & $0.3384 \pm 0.1451$ & $0.5047 \pm 0.1573$ & $0.8817 \pm 0.1842$ & $0.7639 \pm 0.0476$ & $0.8044 \pm 0.1333$ & $0.9903 \pm 0.0300$ \\
MARG-V2-GPT4o         & $0.3077 \pm 0.1400$ & $0.4957 \pm 0.1345$ & $0.8896 \pm 0.1755$ & $0.7763 \pm 0.0714$ & $0.8156 \pm 0.0970$ & $0.9900 \pm 0.0403$ \\
MARG-V2-GPT4o-mini    & $0.2137 \pm 0.1523$ & $0.5167 \pm 0.1397$ & $0.9361 \pm 0.1241$ & $0.7701 \pm 0.0601$ & $0.8644 \pm 0.0643$ & $0.9887 \pm 0.0431$ \\
\midrule
ReviewAgent-3.5-haiku                   & $0.1781 \pm 0.1096$ & $0.6867 \pm 0.1161$ & $0.6868 \pm 0.1825$ & $0.8380 \pm 0.0310$ & $0.6911 \pm 0.1292$ & $\bf{0.9960 \pm 0.0219}$ \\
ReviewAgent-3.7-sonnet                  & $0.4135 \pm 0.1628$ & $0.7213 \pm 0.1048$ & $0.7915 \pm 0.2026$ & $0.8202 \pm 0.0751$ & $0.9222 \pm 0.0823$ & $0.9837 \pm 0.0633$ \\
ReviewAgent-Deepseek-Imp                & $0.2830 \pm 0.1542$ & $0.7313 \pm 0.1205$ & $0.7205 \pm 0.1909$ & $0.7778 \pm 0.1138$ & $0.9356 \pm 0.0866$ & $0.9890 \pm 0.0474$ \\
ReviewAgent-Deepseek-ImpRef             & $0.2927 \pm 0.1253$ & $0.7330 \pm 0.1213$ & $0.7369 \pm 0.1863$ & $0.8021 \pm 0.0676$ & $0.9533 \pm 0.0610$ & $0.9943 \pm 0.0310$ \\
ReviewAgent-Deepseek-Ref                & $0.2883 \pm 0.1192$ & $0.7587 \pm 0.1443$ & $0.7377 \pm 0.1573$ & $0.7767 \pm 0.1057$ & $0.9000 \pm 0.0971$ & $0.9833 \pm 0.0582$ \\
ReviewAgent-GPT4o                       & $0.2740 \pm 0.1592$ & $0.6773 \pm 0.1228$ & $0.7565 \pm 0.1693$ & $0.8018 \pm 0.0954$ & $0.6889 \pm 0.1360$ & $0.9833 \pm 0.0913$ \\
ReviewAgent-GPT4o-mini                  & $0.2029 \pm 0.1391$ & $0.6287 \pm 0.1235$ & $0.7421 \pm 0.1600$ & $0.8092 \pm 0.0906$ & $0.6822 \pm 0.0954$ & $0.9850 \pm 0.0634$ \\
ReviewAgent-Qwen-Imp                     & $0.6469 \pm 0.1415$ & $\bf{0.8787 \pm 0.0990}$ & $0.6649 \pm 0.1727$ & $\bf{0.8401 \pm 0.0347}$ & $0.9822 \pm 0.0300$ & $0.9707 \pm 0.0474$ \\
ReviewAgent-Qwen-ImpRef                  & $\bf{0.6579 \pm 0.1194}$ & $0.8493 \pm 0.1242$ & $0.7418 \pm 0.1848$ & $0.7965 \pm 0.0557$ & $\bf{0.9889 \pm 0.0253}$ & $0.9770 \pm 0.0341$ \\
ReviewAgent-Qwen-Ref                     & $0.6135 \pm 0.1394$ & $0.8347 \pm 0.1058$ & $0.6605 \pm 0.1861$ & $0.8185 \pm 0.0952$ & $0.9778 \pm 0.0535$ & $0.9590 \pm 0.0471$ \\
\midrule
AI-Scientist-3.5-Haiku         & $0.3257 \pm 0.1448$ & $0.5360 \pm 0.0941$ & $0.8624 \pm 0.1441$ & $0.7786 \pm 0.0696$ & $0.6133 \pm 0.1071$ & $0.9953 \pm 0.0256$ \\
AI-Scientist-3.7-Sonnet        & $0.3215 \pm 0.1439$ & $0.7980 \pm 0.1071$ & $0.8208 \pm 0.1694$ & $0.7986 \pm 0.0978$ & $0.9511 \pm 0.0741$ & $0.9667 \pm 0.1269$ \\
AI-Scientist-GPT4o            & $0.3071 \pm 0.1471$ & $0.5217 \pm 0.1100$ & $0.9079 \pm 0.1282$ & $0.7661 \pm 0.0818$ & $0.5911 \pm 0.1144$ & $0.9640 \pm 0.1148$ \\
AI-Scientist-GPT4o-mini       & $0.3406 \pm 0.1390$ & $0.5107 \pm 0.0957$ & $0.8279 \pm 0.1776$ & $0.7947 \pm 0.0417$ & $0.6111 \pm 0.0944$ & $0.9557 \pm 0.1903$ \\
\bottomrule
\end{tabular}
}
\vspace{-0.5em}
\caption{Evaluation of AI-generated reviews across six different metrics on CoreCorpus-30 dataset. ReviewAgent against MARG and Sakana AI Scientist. ReviewEval framework has two binary settings: \textbf{Imp} (improvement decribed in \ref{iterative_loop}) and \textbf{Ref} (reflection described in \ref{relfection_loop}). A suffix of \texttt{-Imp}, \texttt{-Ref} or \texttt{-ImpRef} indicates that the corresponding feature was turned “on”; absence of any suffix implies both features were “off.”}
\label{corecorpus_results}
\end{table*}

\vspace{-0.5em}
\section{Experiments \& Results}\label{sec:evaluation}
\vspace{-0.5em}

\begin{table}[]
\scriptsize
\begin{tabular}{|l|l|l|l|l|l|}
\hline
\multicolumn{1}{|c|}{\textbf{Venue (Year)}} & \multicolumn{1}{c|}{\textbf{Poster}} & \multicolumn{1}{c|}{\textbf{Spotlight}} & \multicolumn{1}{c|}{\textbf{Oral}} & \multicolumn{1}{c|}{\textbf{Rejected}} & \multicolumn{1}{c|}{\textbf{Total}} \\ \hline
UAI (2024)                                  & 8                                    & 8                                       & 8                                  & –                                      & 24                                  \\ \hline
NeurIPS (2024)                              & 8                                    & 8                                       & 8                                  & 8                                      & 32                                  \\ \hline
ICLR (2025)                                 & 8                                    & 8                                       & 8                                  & 8                                      & 32                                  \\ \hline
ICLR (2024)                                 & 8                                    & 8                                       & 8                                  & 8                                      & 32                                  \\ \hline
\textbf{Overall}                            & –                                    & –                                       & –                                  & –                                      & \textbf{120}                        \\ \hline
\end{tabular}
\caption{Composition of our FullCorpus-120 test set.}
\label{tab:dataset-composition}
\vspace{-2em}
\end{table}

\begin{table}[]
\centering
\small
\begin{tabular}{@{}l c@{}}
\toprule
Domain                           & Papers \\
\midrule
Healthcare/Biology               & 11     \\
Social Sciences/Humanities       & 4      \\
Economics/Finance                & 14     \\
Physics/Astronomy                & 6      \\
Engineering/Systems              & 4      \\
\bottomrule
\end{tabular}
\caption{Cross-domain paper count in FullCorpus-120.}
\label{tab:papers-by-domain}
\vspace{-2em}
\end{table}
\label{results-section}
\textbf{Dataset.} \blueadd{We constructed a dataset, \textbf{FullCorpus-120}, comprising 120 papers sampled from the NeurIPS, ICLR, and UAI conferences. For each of the acceptance and rejection category we sampled 8 papers, resulting in a balanced set across acceptance categories (mentioned in table \ref{tab:dataset-composition}). These conferences were selected due to the availability of official reviewer comments for their 2024 and later editions on Openreview.net. Notably, the recent nature of these papers creates a challenging test set, as SOTA LLMs are unlikely to be trained on this data given their data cutoffs dates. This design allows for a robust assessment of each model's ability to interpret and analyze previously unseen data.\\ To promote topical diversity, we select papers that integrate artificial intelligence with other domains such as biology, social science, finance, and engineering (AI + X), thereby avoiding over-representation of any single domain (mentioned in table \ref{tab:papers-by-domain}). We use FullCorpus-120 to establish results for our own reviewer models as well as all baseline methods. \\To evaluate on expensive foundation models and run higher-cost framework settings, we curated a stratified subset comtaining 30 papers \textbf{CoreCorpus-30} that preserves the diversity and balance of the FullCorpus-120 while enabling more efficient experimentation.}\par

\noindent\textbf{Baselines and models.} \blueadd{We compare our AI-generated review approach with two established methods: Sakana AI Scientist~\cite{lu2024ai} and MARG (v1 and v2)~\cite{d2024marg}. To demonstrate the effectiveness of ReviewAgent, we report results using foundational models from Deepseek, Qwen, OpenAI, and Anthropic.}\par

\noindent\textbf{Results for FullCorpus-120.} \blueadd{Table~\ref{tab:all_results} summarizes the performance of ReviewAgent along with baselines and expert reviews on FullCorpus‑120. For each of the six metrics, we highlight the top‑scoring system and contrast it with ReviewAgent:}\par

\noindent\blueadd{\textit{Actionable insights.} MARG‑v1‑qwen narrowly beats  ReviewAgent‑qwen. The fact that ReviewAgent‑qwen nearly matches MARG’s top score demonstrates the efficacy of ReviewAgent in producing concrete, actionable feedback.}\par

\noindent\blueadd{\textit{Adherence to the review guidelines.} Experts average approximately 0.6. ReviewAgent‑Qwen (mean $\approx 0.85$) and ReviewAgent‑DeepSeek (mean $\approx 0.75$) show strong compliance, surpassing both MARG and AI-Scientist, which are hardcoded to a particular review format. In contrast, ReviewAgent dynamically adjusts itself to the given guidelines.}\par

\noindent\blueadd{\textit{Topic coverage and semantic similarity.} AI-Scientist-Deepseek leads the topic coverage, while ReviewEval also shows promising results. For semantic similarity, AI-Scientist‑qwen tops while ReviewAgent‑deepseek closely follows.}

\noindent\blueadd{\textit{Depth of analysis.} ReviewAgent‑qwen achieves the highest depth of analysis score, surpassing both expert reviewers and all baseline models. This demonstrates ReviewAgent’s ability to produce in-depth rather than superficial evaluations.}\par

\noindent\blueadd{\textit{Factual correctness.} MARG‑v2‑deepseek scores the highest, ReviewAgent‑deepseek and ReviewAgent‑qwen remain just below, indicating reliability and low risks of misinterpretations by AI-based frameworks.\\}
\blueadd{Overall, ReviewAgent attains the highest depth, adherence, and near‑top actionable insights, demonstrating its superiority over AI baselines and expert reviews on FullCorpus‑120.}

\noindent\blueadd{\textbf{Results for CoreCorpus-30.} Building on this foundation, we evaluated ReviewAgent under more resource intensive settings and also used proprietary, closed‑source models on CoreCorpus‑30, a stratified subset of FullCorpus‑120. Table \ref{corecorpus_results} summarises these results:}\par

\noindent\blueadd{\textit{Actionable insights.}\\ ReviewAgent‑Qwen‑Improvement‑Reflection leads, closely followed by the Improvement variant. Both outperform all MARG and AI-Scientist baselines, underscoring how feedback from ReviewEval and iterative self‑reflection loop boosts the constructiveness of feedback.}

\noindent\blueadd{\textit{Adherence to the review guidelines.}\\ ReviewAgent‑Qwen‑Improvement achieves the highest adherence, with the Improvement variant a close second followed by Reflection variant. These results exceed all baseline models, including MARG and AI-Scientist, and highlight the effectiveness of dynamic prompt generation and evaluation-driven calibration, coupled with iterative refinements.}

\noindent\blueadd{\textit{Topic coverage and Semantic Similarity.} While the top single-model coverage is achieved by a MARG variant, ReviewAgent‑3.7-sonnet still maintains a strong, balanced scope even if not maximized for breadth.
ReviewAgent‑Qwen-Improvement achieves the highest alignment with expert phrasing. This indicates that ReviewAgent can mimic the tone and delivery of expert comments.}

\noindent\blueadd{\textit{Depth of analysis.} \\ReviewAgent‑Qwen‑Improvement‑Reflection achieves the top score, with the Improvement variant next. Both significantly surpass MARG and AI-Scientist models, validating that iterative reflection enables truly in‑depth, expert‑level critique rather than superficial reviews.}

\noindent\blueadd{\textit{Factual correctness.} ReviewAgent variants achieve among the highest factual correctness scores, with DeepSeek and Haiku-based versions performing best. Other baselines also score high in factual accuracy, showing that the models generally stay close to the source material and avoid making things up. }
\blueadd{Overall, the pronounced gains from the Improvement and Reflection loops demonstrate their effectiveness in elevating review quality.}

\section{Conclusion}
We introduced ReviewEval, a framework evaluating AI-generated reviews on alignment, accuracy, depth, constructiveness, and guideline adherence. \blueadd{Additionally, we propose ReviewAgent, an AI system featuring (1) conference-specific review alignment, (2) iterative self-refinement, and (3) external improvement via ReviewEval. Our experiments show ReviewAgent matches expert quality and surpasses existing AI frameworks.}

\section{Acknowledgement}
The authors wish to acknowledge the use of ChatGPT in improving the presentation and grammar of the paper. The paper remains an accurate representation of the authors’ underlying contributions.

\section{Limitations}
\blueadd {
\textbf{(1)} Although we attempted to enhance cross‑domain diversity by considering AI + X papers, our evaluation dataset comprised papers only from the AI conferences, which may confine the generalizability of our results to other domains, disciplines, or larger‑scale datasets. \textbf{(2)} Dependence on LLMs brings with it familiar weaknesses of LLM-as-a-Judge paradigm: hallucinations, bias, prompt sensitivity, temporal stability and black‑box behavior, limiting interpretability, reliability, transparency, and trust. We tried to overcome this by decomposing the evaluation process itself into a well-defined, multi-step procedure and doing a comprehensive evaluation of our approach across a wide variety of LLMs. \textbf{(3)} Metric interdependence (semantic similarity, factual accuracy, depth of analysis) injects subjectivity and can constrain model flexibility. \textbf{(4)} Iterative refinement loop, external improvement loop and RAG components in our approach have nontrivial computational expense. However, this is shared by many Generative AI tools and products such as Deep Research Agents. Furthermore, we could also incorporate cost-reduction mechanisms like caching, batch processing, and chunk optimisation to enhance efficiency. \textbf{(5)} Lastly, purely automated metrics might not be able to pick up on qualitative aspects like tone, subtle criticism, and readability. 
}
\bibliography{custom}

\appendix

\section{Appendix}
\label{sec:appendix}

\subsection{Correlation Tests of parameters in review eval}
\subsubsection{Averaging Analysis: Metric Impact and Contribution}
\label{sec:averaging_analysis}

To understand the individual importance and contribution of each metric to an overall assessment, an averaging analysis was performed. This analysis aimed to quantify how sensitive a composite score is to the removal of any single metric and to determine the proportional contribution of each metric to this composite score.\\
\textbf{Methodology.} A unified score was first calculated for each data point (representing a specific model's evaluation of a paper) by taking the simple arithmetic mean of the six core metrics: depth score, actionable insights, adherence score, coverage, semantic similarity, and factual correctness.

For each model and for each of the six metrics, the analysis involved the following steps:
\begin{enumerate}
    \item The baseline unified score was established using all six metrics.
    \item Each metric was temporarily removed, and a new adjusted average score was calculated based on the remaining five metrics.
    \item The absolute change and relative percentage change between the baseline unified score and the adjusted score were computed. A positive change indicates that removing the metric increased the overall score (implying the metric typically scored lower than others), while a negative change indicates that removing the metric decreased the overall score (implying the metric typically scored higher).
    \item The contribution of each individual metric to the original unified score was calculated as its value divided by the number of metrics (6). The average percentage contribution of each metric to the baseline unified score across all evaluated papers for a model was also determined.
\end{enumerate}
These calculations were performed for every model in the dataset, and the results were then aggregated to obtain mean and standard deviation values for the changes and contributions across all models.

\textbf{Results.} The aggregated results, presented in Table~\ref{tab:metric_importance_summary}, highlight the overall impact and contribution of each metric. The table shows the mean absolute change in the unified score when a metric is removed, the mean relative percentage change, and the mean percentage contribution of each metric to the unified score.

\begin{table*}[t]         
  \centering
  \begin{adjustbox}{}  
    \begin{tabular}{lrrr}
      \toprule
      Metric Removed & Mean Abs.\ Change & Mean Rel.\ Change (\%) & Mean Contr.\ (\%) \\
      \midrule
      actionable\_insights   &  0.0654 &  9.00  &  9.17 \\
      factual\_correctness   & -0.0492 & -6.96  & 22.47 \\
      depth\_score           & -0.0225 & -2.97  & 19.14 \\
      adherence\_score       &  0.0131 &  1.91  & 15.08 \\
      semantic\_similarity   & -0.0061 & -0.81  & 17.34 \\
      coverage               & -0.0008 & -0.16  & 16.80 \\
      \bottomrule
    \end{tabular}
  \end{adjustbox}
  \caption{Metric Importance Summary: Impact of Removal and Contribution to Unified Score (Averaged Across Models)}
  \label{tab:metric_importance_summary}
\end{table*}

As observed in Table~\ref{tab:metric_importance_summary}, removing actionable insights led to the largest positive mean change in the unified score (0.0654, or +9.00\%), suggesting it generally scored lower than the average of other metrics. Conversely, removing factual correctness resulted in the largest negative mean change (-0.0492, or -6.96\%), indicating its scores were typically higher and thus its removal decreased the overall score most significantly.

In terms of contribution, factual correctness had the highest average contribution to the unified score (22.47\%), followed by depth score (19.14\%) and semantic similarity (17.34\%). Actionable insights had the lowest average contribution (9.17\%).

\begin{figure}[H]
    \centering
    \includegraphics[width=1\linewidth]{ 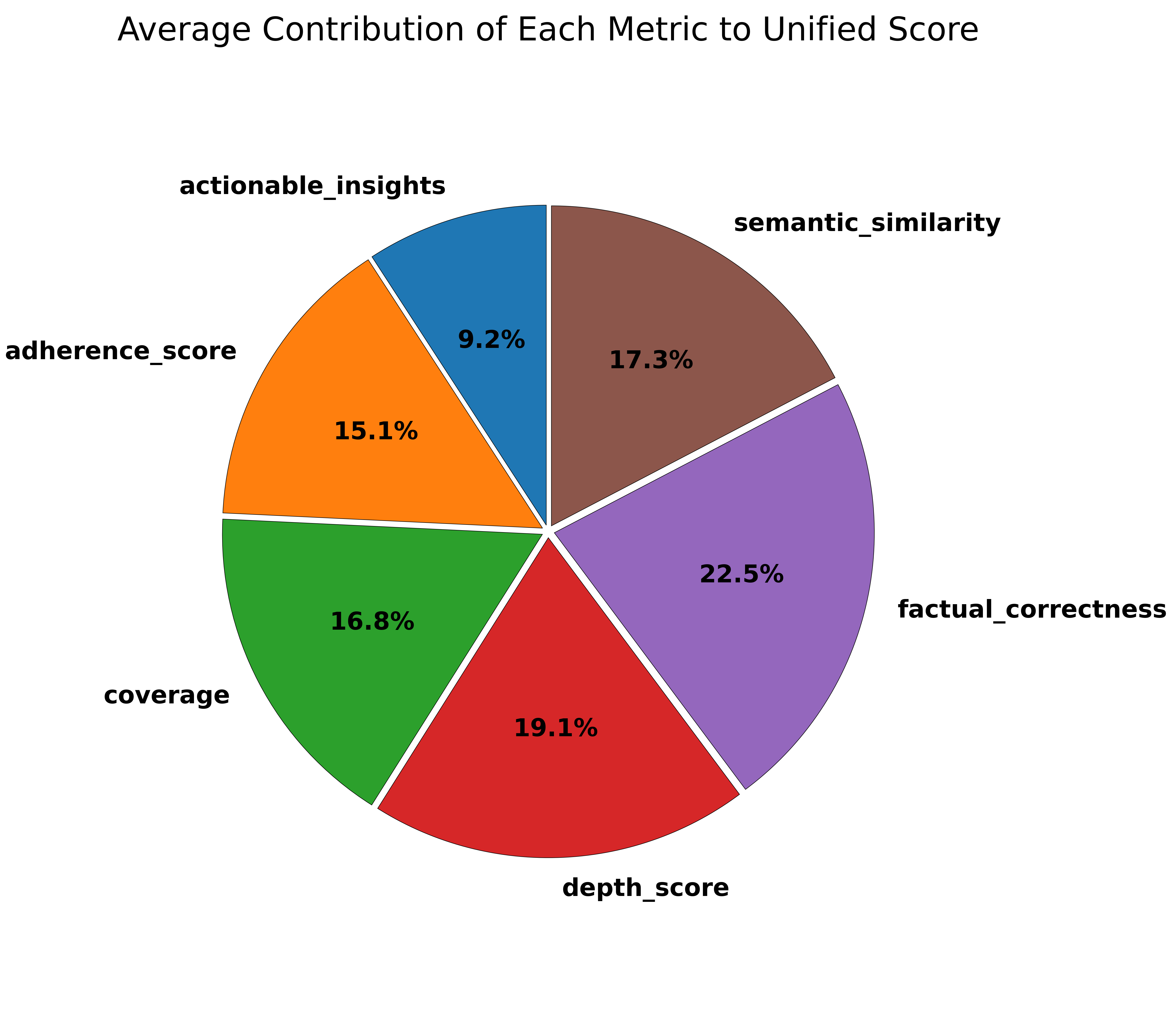}
    \caption{Metric Contribution}
    \label{fig:metric-contri}
\end{figure}

\begin{figure}[H]
    \centering
    \includegraphics[width=1\linewidth]{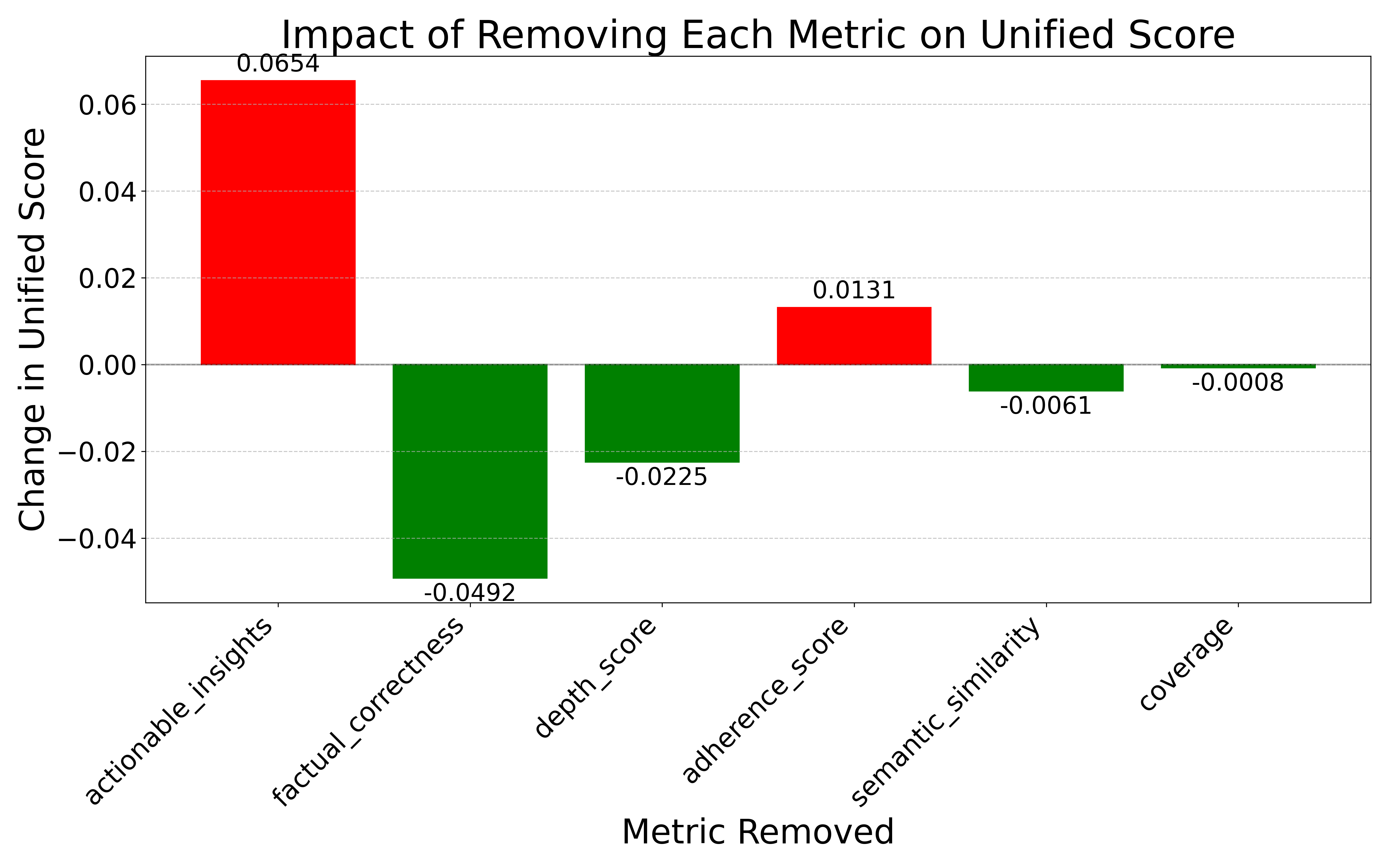}
    \caption{Metric Importance}
    \label{fig:metric-importance}
\end{figure}

These findings are further visualized in the generated plots. Bar chart \ref{fig:metric-importance} illustrates the mean absolute change in the unified score upon removing each metric, with bars colored to distinguish between positive and negative impacts. Pie chart \ref{fig:metric-contri} depicts the average percentage contribution of each metric to the total unified score, providing a clear visual breakdown of metric influence.



\subsubsection{Pearson Correlation Analysis}
\label{sec:pearson_correlation}

To investigate the linear relationships between the different evaluation metrics, a Pearson correlation analysis was conducted. This method assesses the strength and direction of association between pairs of continuous variables. The metrics analyzed were depth score, actionable insights, adherence score, coverage, semantic similarity, and factual correctness.

\textbf{Methodology}
The Pearson correlation coefficient ($r$) was calculated for every pair of the six metrics using all available individual data points. The correlation coefficient ranges from -1 (perfect negative linear correlation) to +1 (perfect positive linear correlation), with 0 indicating no linear correlation. For each correlation, a p-value was also computed to determine the statistical significance of the observed relationship.

\textbf{Results}
The computed Pearson correlation matrix is presented in Table~\ref{tab:correlation_matrix}, and the corresponding p-values are shown in Table~\ref{tab:correlation_pvalues}.

\begin{table*}[htbp]
  \centering
  \begin{adjustbox}{max width=\textwidth}
    \begin{tabular}{lrrrrrr}
      \toprule
       & \texttt{Depth} & \texttt{Actionable Ins.} 
       & \texttt{Adherence Sc.} & \texttt{Coverage} 
       & \texttt{Semantic Sim.} & \texttt{Factual Corr.} \\
      \midrule
      \texttt{Depth Score}         &  1.000 & 0.298 & 0.426 & -0.013 & 0.242 & 0.026 \\
      \texttt{Actionable Insights} &  0.298 & 1.000 & 0.186 & -0.213 & 0.028 & -0.017 \\
      \texttt{Adherence Score}     &  0.426 & 0.186 & 1.000 &  0.030 & 0.203 & -0.001 \\
      \texttt{Coverage}             & -0.013 & -0.213 & 0.030 & 1.000 & 0.155 & -0.015 \\
      \texttt{Semantic Similarity} &  0.242 & 0.028 & 0.203 & 0.155 & 1.000 &  0.045 \\
      \texttt{Factual Correctness} &  0.026 & -0.017 & -0.001 & -0.015 & 0.045 & 1.000 \\
      \bottomrule
    \end{tabular}
  \end{adjustbox}
  \caption{Pearson Correlation Matrix of Metrics}
  \label{tab:correlation_matrix}
\end{table*}

\begin{table*}[htbp]
\centering
\caption{P-values for Pearson Correlations}
\label{tab:correlation_pvalues}
\resizebox{\textwidth}{!}{%
\begin{tabular}{lrrrrrr}
\toprule
 & \texttt{Depth Sc.} & \texttt{Actionable Ins.} & \texttt{Adherence Sc.} & \texttt{Coverage} & \texttt{Semantic Sim.} & \texttt{Factual Corr.} \\
\midrule
\texttt{Depth Score}         & 0.00e+00 & <.001 & <.001 & 0.576 & <.001 & 0.274 \\
\texttt{Actionable Insights} & <.001 & 0.00e+00 & <.001 & <.001 & 0.225 & 0.479 \\
\texttt{Adherence Score}    & <.001 & <.001 & 0.00e+00 & 0.191 & <.001 & 0.952 \\
\texttt{Coverage}            & 0.576 & <.001 & 0.191 & 0.00e+00 & <.001 & 0.510 \\
\texttt{Semantic Similarity} & <.001 & 0.225 & <.001 & <.001 & 0.00e+00 & 0.055 \\
\texttt{Factual Correctness} & 0.274 & 0.479 & 0.952 & 0.510 & 0.055 & 0.00e+00 \\
\bottomrule
\end{tabular}%
}
\end{table*}

The strongest positive correlation was observed between \texttt{Adherence Score} and \texttt{Depth Score} ($r \approx 0.426$, $p < 0.001$), suggesting that evaluations with higher adherence to instructions also tended to have greater depth. Another notable positive correlation was found between \texttt{Depth Score} and \texttt{Actionable Insights} ($r \approx 0.298$, $p < 0.001$).
\texttt{Coverage} showed a statistically significant negative correlation with \texttt{Actionable Insights}, indicating that broader coverage might sometimes come at the expense of providing actionable insights, or vice-versa.
Most other correlations were relatively weak, although several were statistically significant due to the large sample size. For instance, the correlation between \texttt{Semantic Similarity} and \texttt{Factual Correctness} was very weak and borderline significant.
\texttt{Factual Correctness} showed very weak and non-significant correlations with the other 5 parameters.

\begin{figure}[H]
    \centering
    \includegraphics[width=1\linewidth]{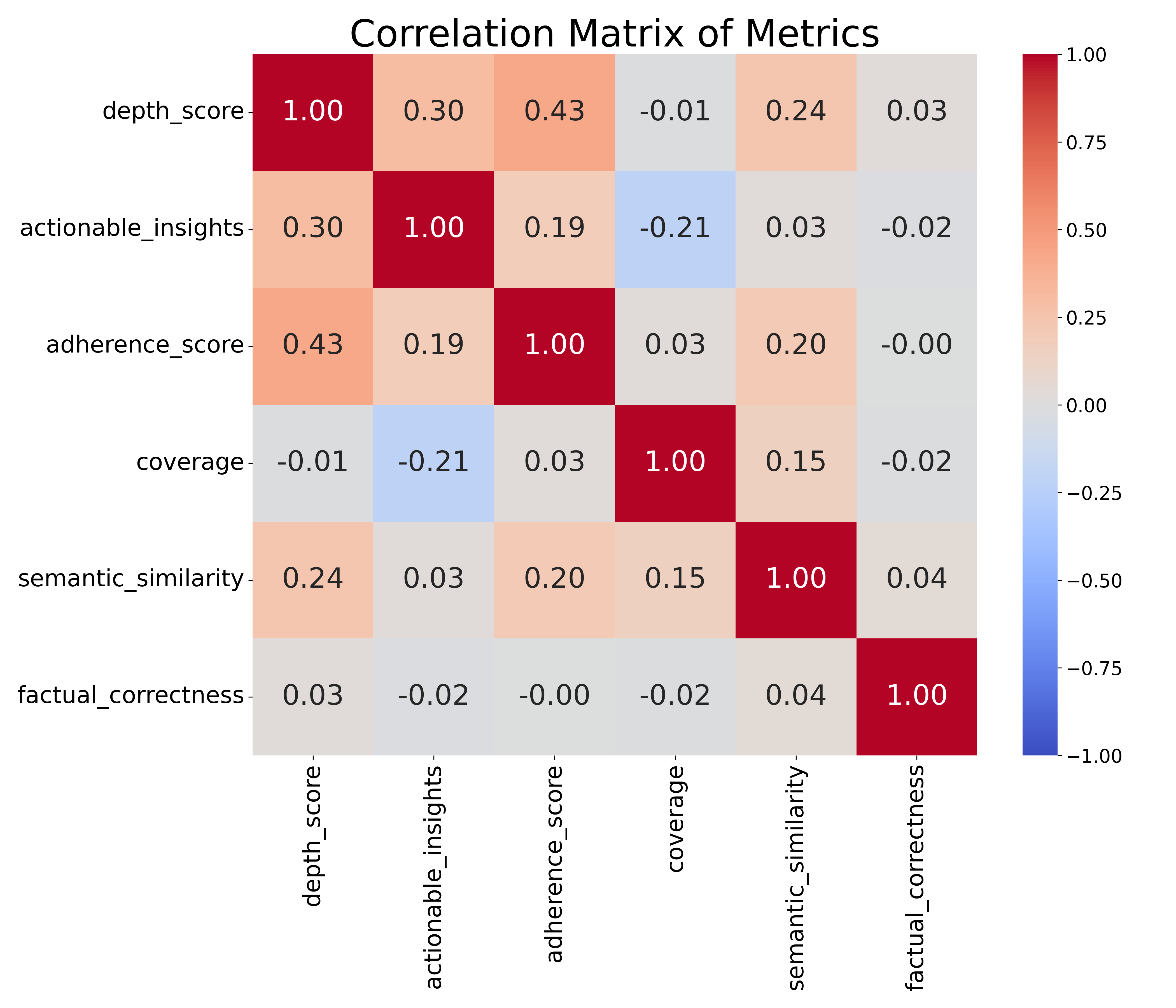}
    \caption{Correlation Heatmap}
    \label{fig:correlation-heatmap}
\end{figure}

\begin{figure}[H]
    \centering
    \includegraphics[width=1\linewidth]{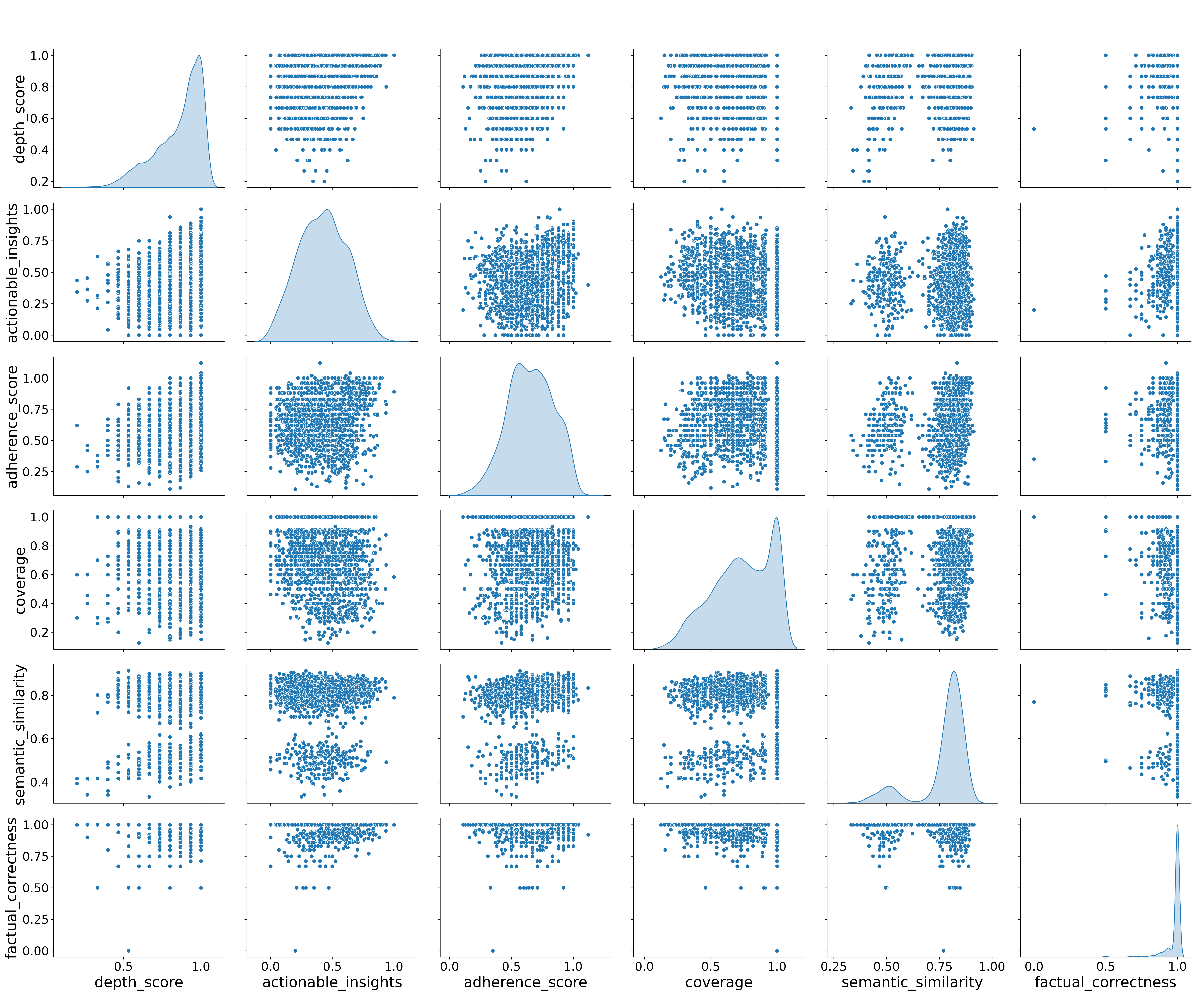}
    \caption{Metric-wise scatter plot}
    \label{fig:metric-scatter-plot}
\end{figure}

These relationships are visually summarized in a heatmap \ref{fig:correlation-heatmap} and a series of scatter plots \ref{fig:metric-scatter-plot}.


\subsection{Examples of question generation and decomposition in factual correctness pipeline}
\label{factual_correctness_examples}
 \textbf{Example of question generation from review point in step 1:} Consider the following review from the PeerRead dataset \cite{kang2018dataset} for the paper \textit{"Augmenting Negative Representations for Continual Self-Supervised Learning"} \cite{chaaugmenting}:

\underline{Review (\(R\)):}  
\texttt{“Augmentation represents a crucial area of exploration in self-supervised learning. Given that the authors classify their method as a form of augmentation, it becomes essential to engage in comparisons and discussions with existing augmentation methods.”}

\underline{Generated question (\(Q\)):}  
\texttt{“Has the paper engaged in comparisons and discussions with existing augmentation methods, given that the authors classify their method as a form of augmentation?”}

\textbf{Example of question decomposition in step 2:}
\underline{Implementation Detail:} We used Langchain’s built-in query decomposition tool for initial question breakdown. Langchain’s query decomposition module leverages large language models (LLMs) to break down complex, multi-part queries into simpler, sequential sub-questions. It typically works by first identifying logical or temporal dependencies within the input question, and then restructuring it into smaller, more manageable components. These sub-questions can then be answered independently or in a reasoning chain. We answer these questions independently using our custom RAG pipeline. \\
\underline{Main Question:} \texttt{Has the paper engaged in comparisons and discussions with existing augmentation methods, given that the authors classify their method as a form of augmentation?\\}
\underline{Sub‐questions:}\\
\texttt{a) Has the paper compared its augmentation method against existing augmentation methods?\\
b) Does the paper discuss the strengths and weaknesses of related augmentation techniques?}

\subsection{Examples of three dimensions of constructiveness}
\label{constructiveness_example}
\begin{itemize}
\item Specificity: \begin{itemize}
        \item \textit{Example (score 1)}: ``Section 5.3 discusses pretraining dataset selection but does not address the potential privacy costs of using private data for this purpose. Refer to \textit{<Research Paper Citation>} for methods to ensure privacy in this step.''
        \item \textit{Non-example (score 0)}: ``The paper lacks novelty and is a straightforward application of existing techniques.''
\end{itemize}

\item Feasibility: \begin{itemize}
        \item \textit{Example (score 1)}: ``Break down the GPU hours into pretraining and fine-tuning stages in Table 7 to make the computational cost more transparent.''
        \item \textit{Non-example (score 0)}: ``Add experiments with a wide variety of datasets, including proprietary and restricted-access data, to generalize findings.''
\end{itemize}
\item Implementation Details: \begin{itemize}
        \item \textit{Example (score 1)}: ``In Algorithm 1, correct the noise addition formula to $1/B \cdot \mathcal{N}(0, \sigma^2 C^2 I)$, as this ensures proper scaling of noise with batch size.''
        \item \textit{Non-example (score 0)}: ``Use advanced techniques to improve DP-SGD.''
\end{itemize}
\end{itemize}

\subsection{Prompts Used}
\label{prompts-used}
Below are the system instructions and prompts employed in our review generation pipeline. These prompts guide each stage of the process—from extracting reviewer guidelines from HTML content, to generating detailed review prompts for specific paper sections, and finally formatting the reviews in strict adherence to conference guidelines.

\textbf{Guidelines Parsing Prompt.} \texttt{You are a smart AI designed to extract reviewer guidelines from HTML content, regardless of its structure or format. You will be provided with the raw HTML of a webpage that contains the guidelines. Your task is to intelligently parse and extract the most relevant content based on the following high-level objectives:
\begin{enumerate}
    \item Understand the Context: The HTML file may contain multiple sections of a webpage, including irrelevant information like headers, footers, navigation bars, or ads. Your goal is to focus solely on extracting meaningful content that pertains to reviewer guidelines. Look for terms such as 'reviewer', 'guidelines', 'evaluation', 'criteria', 'instructions', or 'review process' that may indicate sections of interest.
    \item Text Structure: Look for relevant sections by identifying common phrases or paragraphs that may contain instructions or rules for reviewers. This includes but is not limited to guidelines on evaluation, reviewing criteria. Focus on only the main content that provides the guidelines for how to review the papers content and not the conference details.
    \item Avoid Noise: Ignore or discard text that is likely irrelevant, such as menus, links to other pages, copyright information, or promotional content. You are interested only in extracting text that provides guidance to reviewers for evaluating papers.
    \item Identify Sections Based on Common Words: You can identify the main sections of interest by finding phrases like:
        ``Reviewer Guidelines'', 
        ``Review Criteria'',
        ``Evaluation Process'',
        ``Instructions for Reviewers'',
        ``Review Process Overview'',
    When you find such phrases, capture the paragraph or section following the phrase, as this is likely to contain the reviewer guidelines.
    \item Extract Text Around These Keywords: When you identify these keywords, extract approximately 3-4 paragraphs surrounding these keywords to capture the guidelines. This includes headings or bullet points that may be present.
    \item Return Results as String: Once you have completed parsing the HTML content and extracted relevant guidelines, return the guidelines as a single continuous string. Ensure the text is well-formatted and readable, without HTML tags or irrelevant information like advertisements or links.
    \item Avoid capturing details of the conference or event itself, such as times, dates, locations, or registration information. Your task is to focus solely on the reviewer guidelines and evaluation criteria.
    \item Avoid capturing details of what software or tools to use for the review process. Focus on the guidelines for evaluating the content of the papers.
    \item  If there is a table of any sort in the reviewer guidelines, extract the text content of the table and present it in a readable format, as a paragraph or list of items. Do not include the table structure in the extracted text.
    \item  if there are guidelines for multiple types of papers like ones in CER OR PCI OR NLP, extract the information of the first type of paper only. Do not give any recitations of any sort as that is blocked by google because of copyright issues. Note that you MUST also check the format which is required by the conference guidelines for a review and the output should be given in that format in the end
\end{enumerate}
}

\textbf{Prompt for instruction generation for review of a section.} \texttt{You are a generative language model (LLM X) creating a prompt for another research paper reviewer LLM (LLM Y), generate a detailed prompt instructing LLM Y on how to review the {section} section of a research paper. Consider the following criteria:
\begin{enumerate}
    \item The clarity and completeness of the {section}.
    \item The relevance and alignment of the {section} with the main themes and objectives of the paper.
    \item The logical consistency and evidence support in the {section}.
    \item The originality and contribution of the {section} to the field.
    \item Any specific elements highlighted in the conference guidelines that should be focused on in the {section}.
\end{enumerate}
Provide structured and clear instructions in the form of a plan with steps that will enable LLM Y to conduct a thorough and critical review of the research paper's {section}. Use the given conference guidelines. Do not give any recitations of any sort as that is blocked by google because of copyright issues.}

\textbf{Prompt for finally formatting review as per conference guidelines.} \texttt{You are an expert in writing reviews for various research paper conferences. You will be given reviews for various sections of a research paper, and the research paper itself and you are supposed to write the review in the format that is expected for submission to the specified conference. You're given the contents of the reviewer guidelines for the conference and you are supposed to adhere to it strictly. You are also not supposed to change the content of the review provided to you AT ALL. You are just a formatter and are supposed to just rewrite the given review into the given format while making the necessary changes. You are to give the complete review of the paper in the format of the conference (the entire paper, not some part of it). Remember that you have an outut token limit of 8192 tokens and your entire review is supposed to fit within that limit, so be careful. This is the conference guidelines for the conference : {str(guidelines)}}

\subsection{Example Reviews}

\noindent\textbf{Below is an example review of a paper which has been accepted by ReviewAgent.}\\ \\ \\

\noindent\textbf{Paper Title:} \textit{Fast and Unified Path Gradient Estimators for Normalizing Flows} (\href{https://arxiv.org/pdf/2403.15881}{arXiv:2403.15881})

\vspace{1em}
\noindent\textbf{Summary}\\
This paper proposes a computationally efficient and unified framework for estimating path gradients in normalizing flows (NFs), enabling scalable training for both reverse and forward KL divergences. The authors address critical inefficiencies in prior path gradient estimation—such as repeated forward–backward passes and costly numerical inversion—via recursive gradient equations and implicit differentiation. Their approach achieves 1.3–8× runtime improvements and higher effective sample size (ESS). The method unifies coupling flows, continuous flows, and implicitly invertible flows, with applications in physics (e.g., U(1) lattice theory) and high-dimensional ML.

\vspace{0.5em}
\noindent\textbf{Strengths}
\begin{itemize}[left=0pt]
  \item \textbf{Clear Problem Statement \& Motivation:} Identifies the 4–5× runtime overhead of existing path gradients as a key barrier for large-scale NF training. Links computational bottlenecks (e.g., numerical inversion) to practical challenges in scalable inference for physics and machine learning.
  \item \textbf{Theoretical \& Methodological Novelty:}
    \begin{itemize}
      \item Derives recursive gradient equations (Corollary 3.4) for affine coupling flows, reducing runtime by avoiding Jacobian inversion.
      \item Reformulates forward KL training as reverse KL in base space (Proposition 4.1), enabling direct application of path gradients with regularization benefits.
      \item Unifies path gradient computation across diverse architectures (coupling, continuous, implicitly invertible).
    \end{itemize}
  \item \textbf{Empirical Validation:} Demonstrates significant ESS improvements (97.4\% for GMMs) and runtime gains (17.5× faster for implicit flows), with comparisons to prior work (e.g., Vaitl et al.\ 2022a).
  \item \textbf{Practical Relevance:} Highlights applications for physics simulations (e.g., $\phi^4$ lattice theory) and high-dimensional problems.
  \item \textbf{Strong Contribution:} The recursive framework and KL duality argument are novel and theoretically rich.
\end{itemize}

\vspace{0.5em}
\noindent\textbf{Weaknesses}
\begin{itemize}[left=0pt]
  \item \textbf{Overspecialized Language:} Uses terms like \emph{pullback density}, \emph{diffeomorphisms}, and \emph{KL duality} without intuitive explanations.
  \item \textbf{Evidence Integration:} Empirical results are emphasized but not contextualized in the introduction/body.
  \item \textbf{Missing Statistical Rigor:}
    \begin{itemize}
      \item Runtime claims lack significance tests.
      \item Missing error bars for ESS in several tables.
      \item Confounding variables (e.g., batch size impact) are not discussed.
    \end{itemize}
  \item \textbf{Domain Restrictions:} Excludes autoregressive flows without quantifying impact, and relies on explicit energy functions limiting broader applicability.
\end{itemize}

\vspace{0.5em}
\noindent\textbf{Soundness}\\
Technical correctness of the recursive gradient computation and implicit differentiation is well established; empirical claims are supported by ESS benchmarks and runtime comparisons, though limitations on non-physical tasks remain unaddressed.

\vspace{0.5em}
\noindent\textbf{Presentation}\\
Accessible to experts but opaque to general readers; figures/tables need annotations and reproducibility details (e.g., optimizer, batch size, hardware).

\vspace{0.5em}
\noindent\textbf{Contribution}\\
Methodological advancement for efficient path gradients in both forward and reverse KL training; novel unification via base-space pullback, with potential for fairness- and physics-based applications.

\vspace{0.5em}
\noindent\textbf{Critical Questions for Authors}
\begin{enumerate}[left=0pt]
  \item Does Proposition 3.3 provide variance reduction beyond runtime?
  \item How adapt to tasks with intractable energy functions?
  \item Which architectures are incompatible with the framework?
  \item Why are certain ESS values missing in Table 1?
\end{enumerate}

\vspace{0.5em}
\noindent\textbf{Recommendation:} \textit{Accept} with revisions.

\vspace{0.5em}
\noindent\textbf{Constructive Feedback}
\begin{enumerate}[left=0pt]
  \item Expand literature comparison to control variate methods.
  \item Clarify KL duality with analogies and roadmaps.
  \item Improve statistical reporting (confidence intervals, significance tests).
  \item Discuss domain limitations explicitly.
  \item Address runtime–tolerance trade-offs with figures.
\end{enumerate}

\vspace{0.5em}
\noindent\textbf{Final Comments}\\
A computationally and theoretically impactful solution to NF path gradient inefficiencies; with enhanced comparisons and statistical rigor, it will be a strong ICLR contribution.

\noindent\textbf{Reviewer Confidence:} High\\
\textbf{Rating:} Accept

\newpage

\noindent\textbf{Below is an example review of a paper which has been rejected by ReviewAgent.}\\ \\ \\
\noindent\textbf{AlloNet Review}\\
\noindent\textbf{Paper Title:} \textit{Allostatic Control of Persistent States in Spiking Neural Networks for perception and computation} (\href{https://arxiv.org/pdf/2503.16085v1}{arXiv:2503.16085v1})

\vspace{1em}
\noindent\textbf{Summary}\\
The paper proposes \emph{AlloNet}, a spiking neural network integrating allostasis (via the Hammel model) with ring attractor dynamics for controlling persistent neural activity (bumps) in numerical cognition tasks like subitization. The model aims to align internal representations with environmental inputs and demonstrates subitization performance consistent with human behavioral trends (e.g., subitizing limits for numbers >3). However, model reaction times are slower than biological benchmarks, and reproducibility details are minimal.

\vspace{0.5em}
\noindent\textbf{Strengths}
\begin{itemize}[left=0pt]
  \item \textbf{Conceptual Integration:} Combines allostasis and ring attractors to control persistent states, offering a biologically motivated framework for abstract cognition.
  \item \textbf{Technical Details:} Provides rigorous mathematical formulations (e.g., synaptic weight equations in Eq.\ 1) and empirical validation through subitization experiments, including reaction time and error rate analysis.
  \item \textbf{Biological Grounding:} Ties the model to hippocampal/entorhinal dynamics and human subitization behavior (Dehaene \& Cohen 1994; Togoli \& Arrighi 2021).
  \item \textbf{Task Relevance:} Subitization experiments capture qualitative aspects of human performance (e.g., numerosity-dependent reaction time variability).
\end{itemize}

\vspace{0.5em}
\noindent\textbf{Weaknesses}
\begin{itemize}[left=0pt]
  \item \textbf{Novelty Overstatements:} Claims a “novel unified framework” without addressing recent work on dynamic ring attractor control (e.g., Khona \& Fiete 2022; Rapu \& Ganguli 2024). The allostatic‑coupling mechanism lacks clear distinction from gain modulation approaches.
  \item \textbf{Reproducibility Deficiencies:} References NEST simulations with hardcoded parameters (e.g., $\tau_1=75\,$ms) but omits implementation details (Poisson spike calibration, hyperparameter search) and public code/data.
  \item \textbf{Limited Theoretical Depth:} Key design choices (e.g., $\sigma_1/\sigma_2$ values, Gaussian synaptic weights in Eq.\ 1) are not justified; no comparison to alternative regulation mechanisms (e.g., reinforcement learning).
  \item \textbf{Narrow Task Scope:} Subitization is the sole task demonstrated; broader claims (e.g., robotics/spatial navigation) are unsubstantiated.
\end{itemize}

\vspace{0.5em}
\noindent\textbf{Soundness}\\
Technical validity of the model dynamics is clear; however, experimental rigor is weakened by missing error bars and statistical comparisons to human benchmarks.

\vspace{0.5em}
\noindent\textbf{Presentation}\\
Jargon (e.g., “HGM/LGM,” “ring attractor”) needs definitions; figures lack error bars and clear labels; the manuscript lacks a dedicated “Contribution” section.

\vspace{0.5em}
\noindent\textbf{Contribution}\\
Advances biologically inspired control of persistent states but reuses established components (Hammel model, ring attractors) without sufficient novelty justification; broader applicability claims are speculative.

\vspace{0.5em}
\noindent\textbf{Questions for Authors}
\begin{enumerate}[left=0pt]
  \item How does AlloNet’s allostatic framework differ from predictive coding approaches in Rapu \& Ganguli (2024) or active inference models (Parr et al.\ 2022)?
  \item What evidence supports “predictive adaptation” compared to non‑learning SNN models?
  \item Why were Gaussian synaptic weights chosen (Eq.\ 1)? Are there theoretical or empirical motivations?
  \item Does the model replicate quantitative error rates from human subitization (e.g., Castaldi et al.\ 2021)?
\end{enumerate}

\vspace{0.5em}
\noindent\textbf{Limitations}\\
Recent SNN literature gaps (e.g., Rapu \& Ganguli 2024) ignored; reaction times (1000 ms) lag biological norms; reliance on manual parameter tuning limits scalability.

\vspace{0.5em}
\noindent\textbf{Ethics Review}\\
No major ethical issues, but authors should discuss potential biases in AI systems mimicking human cognitive processes.

\vspace{0.5em}
\noindent\textbf{Rating \& Confidence}\\
\textbf{Rating:} 5/10 (technically sound but under‑theorized and narrow)\\
\textbf{Recommendation:} \emph{Reject Unless Major Revisions Address} (novelty justification, reproducibility materials)\\
\textbf{Confidence:} High

\vspace{0.5em}
\noindent\textbf{Additional Comments}
\begin{enumerate}[left=0pt]
  \item Add a dedicated “Contribution” section post‑introduction.
  \item Contrast with recent work (e.g., Rapu \& Ganguli 2024; Rapp \& Nawrot 2020).
  \item Release code and hyperparameters on a public repository.
  \item Demonstrate broader tasks (e.g., spatial navigation) for claimed applicability.
\end{enumerate}

\vspace{0.5em}
\noindent\textbf{Final Outcome:} \emph{Reject with Major Revisions}\\
\textbf{Confidence in Decision:} High

\end{document}